\documentclass[conference]{IEEEtran}
\ifCLASSINFOpdf
\else
\fi
\usepackage{amsmath}
\usepackage{array}
\usepackage{lipsum}
\usepackage{siunitx}
\usepackage{amssymb}
\usepackage{tikz}
\usetikzlibrary{calc,patterns,decorations.pathmorphing,decorations.markings,positioning,backgrounds,arrows.meta,shapes,fit,matrix}
\usepackage{float}

\begin{document}
%
\title{Elasticity Measurements of Expanded Foams \\ using a Collaborative Robotic Arm
}

\author{\IEEEauthorblockN{Luca Beber, Edoardo Lamon, Luigi Palopoli}
\IEEEauthorblockA{Dept. of Eng. and Computer Science\\
University of Trento\\
Trento, Italy\\
Email: \textit{{name.surname}@unitn.it}}
\and
\IEEEauthorblockN{Luca Fambri, Matteo Saveriano, Daniele Fontanelli}
\IEEEauthorblockA{Dept. of Industrial Engineering\\
University of Trento\\
Trento, Italy\\
Email: \textit{{name.surname}@unitn.it}}
}


%


\maketitle
\textit{This work has been submitted to the IEEE for possible publication. Copyright may be transferred without notice, after which this version may no longer be accessible.}

$\\$ 

\begin{abstract}
Medical applications of robots are increasingly popular to objectivise and speed up the execution of several types of diagnostic and therapeutic interventions. Particularly important is a class of diagnostic activities that require physical contact between the robotic tool and the human body, such as palpation examinations and ultrasound scans. The practical application of these techniques can greatly benefit from an accurate estimation of the biomechanical properties of the patient's tissues. In this paper, we evaluate the accuracy and precision of a robotic device used for medical purposes in estimating the elastic parameters of different materials. The measurements are evaluated against a ground truth consisting of a set of expanded foam specimens with different elasticity that are characterised using a high-precision device. The experimental results in terms of precision are comparable with the ground truth and suggest future ambitious developments.
\end{abstract}


%
\IEEEpeerreviewmaketitle

\section{Introduction}

The defining feature of medical robots is that to carry out their
tasks properly they have to come into physical contact with the human
body.  This simple fact emphasises the importance of safety
constraints in the interaction phase.
To meet such stringent requirements, robots need contact models to estimate the extension of the contact area, estimate the interaction forces and the end-effector penetration into the external layers of the body. 
In this context, very popular is the Hunt-Crossley contact
model~\cite{Zhu2021ExtendedModel,haddadi2012real,Pappalardo2016Hunt-CrossleySurgery},
which takes into account the non-linear behaviour of the forces
resulting from the three-dimensional nature of the contact. Despite
its proven effectiveness, the model requires a difficult parameters
identification phase (usually based on the solution of complex
optimisation problems) and it does not allow
establishing a clear connection between the model stiffness coefficient and the material elasticity. This is a major drawback for us, since elasticity, and more in general, visco-elasticity is known to play an important role in medical applications. For instance, the presence of a stiff area in the abdominal region may indicate the presence of a cancerous node~\cite{Greenleaf2003SelectedTissues}. Other similar exams can reveal the presence of systemic sclerosis (SSC)~\cite{Dobrev1999InSclerosis} or can be used to characterise the ageing phenomena~\cite{Lau2008IndentationCartilage}.

Over the years, various techniques have been established to quantify the visco-elastic properties of the human body. Magnetic resonance tissue viscoelasticity, for example, estimates these parameters by assessing the shear wave propagation velocity~\cite{Tang2015UltrasoundTechniques,Manduca2021MRTerminology}. Ultrasound elastography utilises the captured images to estimate visco-elasticity by following the propagation of waves~\cite{Frulio2013UltrasoundLiver,Li2017MechanicsElastography,Tang2015UltrasoundTechniques,Kumar2010MeasurementApplications,Eder2007PerformanceResolution}. These techniques are highly efficient but require specific instrumentation to derive accurate estimates, and are in general too uncertain to collect measurements about the contact forces between the robot and the human body.

For these reasons, we have adopted the established method of \textit{Dimensionality Reduction}~\cite{Popov2015MethodFriction}, commonly employed in tribology, to reconstruct contact forces and determine the elasticity value. 
One of the most appealing features of the method is its ability to expose 
the correlation between the stiffness coefficient, the end-effector shape and the shape of the tip. This way, it is possible to estimate the elasticity values using instruments with more complex shapes than the classic spherical shape, for which the Hertzian theory can be used to estimate force and elasticity~\cite{johnson1985contact}.

\begin{figure}[t]
  \centering
  \input{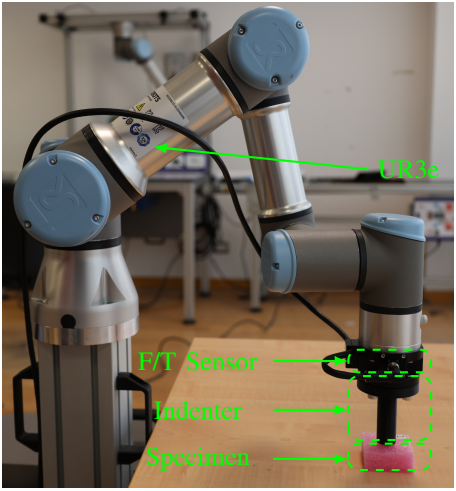} 
  \caption{Experimental setup with the robotic arm made of a 6 DoF Ur3e robotic arm, a 6-axis F/T sensor, a 3D printed indenter and the speciment that is been tested.}
  \label{fig:robot_photo}
\end{figure}

Since our final goal is to enable palpation or ultrasound scans using a remote or autonomous robot station, we need an 
appropriate instrument embedded in the robotic arm to 
collect the measurement and estimate the requested parameters within 
uncertainty. Our choice has fallen on a 6-axis force-torque sensor,
which is applied on an end-effector consisting of a 3D-printed
indenter, which can be used to execute the palpation tests. The whole
system can be seen in Figure~\ref{fig:robot_photo}.

Whilst static tests are sufficient to estimate the elasticity
value~\cite{Hayes1972ACartilage,Sakamoto1996ATest,
  Waters1965TheIndentors, Dimitriadis2002DeterminationMicroscope}, the
primary challenge with this estimation is determining the precise
location of the surface in order to calculate penetration,
particularly when working in unknown environments and when using a
force sensor with unavoidable uncertainties. In this paper, we
propose an estimation algorithm that addresses this important
limitations of the literature.  The final goal of the algorithm is to
estimate the contact force between the end-effector and the surface
tissues of the human body along with the elastic coefficient of the
latter. The paper presents the main ideas underlying the described
process and the results of the characterisation of the measurement
system using different classes of materials with known model
parameters. In particular, three expanded foams are used as test materials, as they show comparable elasticity values compared to biological tissues~\cite{wells2011medical}. 
In addition, the position of the end effector and the
contact forces were acquired to estimate the elasticity coefficient,
taking into consideration the shape of the end-effector probing
tip. At the same time, we determine the position of the surface that
minimises the residuals of a least squares sum, which optimally
calculates the elasticity value.
Our findings show that the level of accuracy reached by the solution is sufficient for its application to the bio-medical context in a near future.



\section{System Models}
\label{sec:models}


A very effective way to describe the contact behaviour of any axial
symmetric indenter on an elastic half-space is through the use of the
Dimensionality Reduction~\cite{Popov2015MethodFriction}. The main idea behind this method is that the 3D contact
between an indenter of an arbitrary shape and the surface of an object
can be reproduced by a one-dimensional linearly elastic foundation,
i.e., the material can be modelled as a set of identical springs
positioned at a small distance. The resulting contact
stiffness is a function of the size of the indenter, of the stiffness of the springs and their distance, i.e.,
\begin{equation}
  \Delta k_z = E^* \Delta x ,
  \label{eq:dr2}
\end{equation} 
with $\Delta k_z$ being the stiffness of a single spring, $\Delta x$
the spacing between the springs and $E^*$ the effective modulus of
elasticity.

The indenter can be considered infinitely stiffer than the
surface, so the elasticity can be extracted knowing the Poisson ratio
$\nu$ as
\begin{equation}
    E_f = E^* (1-\nu^2) .
    \label{eq:elast}
\end{equation}
The force exerted by each spring is expressed as a function of the
deformation using~\eqref{eq:dr2} and \eqref{eq:elast}, that is
\begin{equation}
    f_{N,i} = \frac{E_f}{1-\nu^2} \Delta x u_{z,i}, 
    \label{eq:single_spring}
\end{equation}
where $u_{z,i}$ is the deformation parameter of the $i$-th spring. The
penetration depth of the indenter is a function of its shape and of
the position of the tip of the end-effector. For example, for a flat
indenter, the springs deformation will be equal everywhere, whereas
with a spherical indenter the spring in the centre of the tip will be
more compressed than the ones on the sides.

\begin{figure}[t]
\centering
    \usetikzlibrary{matrix,fit} 
\usetikzlibrary{calc,patterns,decorations.pathmorphing,decorations.markings,positioning,backgrounds,arrows.meta,shapes,fit,matrix}
\begin{tikzpicture}[every node/.style={outer sep=0pt},thick,
 mass/.style={draw,thick},
 spring/.style={thick,decorate,decoration={zigzag,pre length=0.3cm,post
 length=0.3cm,segment length=6}},
 spring2/.style={thick,decorate,decoration={zigzag,pre length=0.3cm,post
 length=0.3cm,segment length=6, angle=10}},
 ground/.style={fill,pattern=north east lines,draw=none,minimum
 width=0.75cm,minimum height=0.3cm},
 dampic/.pic={\fill[white] (-0.1,-0.3) rectangle (0.3,0.3);
 \draw (-0.3,0.3) -| (0.3,-0.3) -- (-0.3,-0.3);
 \draw[line width=1mm] (-0.1,-0.3) -- (-0.1,0.3);}]

  \tikzstyle{damper}=[thick,decoration={markings,  
  mark connection node=dmp,
  mark=at position 0.5 with 
  {
    \node (dmp) [thick,inner sep=0pt,transform shape,rotate=-90,minimum width=15pt,minimum height=3pt,draw=none] {};
    \draw [thick] ($(dmp.north east)+(2pt,0)$) -- (dmp.south east) -- (dmp.south west) -- ($(dmp.north west)+(2pt,0)$);
    \draw [thick] ($(dmp.north)+(0,-5pt)$) -- ($(dmp.north)+(0,5pt)$);
  }}, decorate]

  \node[ground, minimum width=5.5cm, minimum height=3mm, anchor=north] (g1){};
  \node[above=1.5cm of g1, draw, circle, minimum size=3cm, anchor=south,fill=lightgray,thick] (cir){};
  \draw[line width = 0.5mm] (cir.-42) -- ++ (1.6,0) coordinate[right] (cr);
  \draw[thin,dashed] (cir.-42) -- ++ (-1.1,0) coordinate[midway] node[midway,above]{$a$};
  \draw[thin] (cir.center) -- (cir.134) coordinate[midway] node[midway,below]{$R_1$};
  \draw[line width = 0.5mm] (cir.222) -- ++ (-1.6,0) coordinate[midway](cl);
  \draw (g1.north west) -- (g1.north east);
  \draw[thin,dashed] (cr) -- ++ (0.75,0) coordinate[midway](ud);
  \draw[thin,dashed] (cr |- cir.south) -- ++ (0.75,0) coordinate[midway](ld);
  \draw[latex-] (ld) -- (ud) node[midway,right]{$d$};
  \draw[line width = 0.5mm] (cir.222) arc (222:318:1.5cm);
  \node[mass,above=5.5cm of g1,minimum width=5cm,minimum height=0.75cm,fill=red!70] (ft_sensor) {$\text{F/T}_\text{sensor}$};
  \draw[thick] (cir.0) -- (ft_sensor.east |- ft_sensor.south) coordinate[midway](l1); 
  \draw[thick] (cir.180) -- (ft_sensor.west |- ft_sensor.south) coordinate[midway](l2); 
        
    \draw[spring] ([xshift=0mm]g1.north) coordinate(aux) 
    -- (aux|-cir.south) node[midway,right=1mm]{};
    \draw[spring2] ([xshift=5mm]g1.north) coordinate(aux)
   -- (aux|-cir.290) node[midway,right=1mm]{};
   \draw[spring] ([xshift=10mm]g1.north) coordinate(aux)
   -- (aux|-cir.312) node[midway,right=1mm]{};
   \draw[spring] ([xshift=15mm]g1.north) coordinate(aux)
   -- (aux|-cr.south) node[midway,right=1mm]{};
   \draw[spring] ([xshift=20mm]g1.north) coordinate(aux)
   -- (aux|-cr.south) node[midway,right=1mm]{};
   \draw[spring] ([xshift=25mm]g1.north) coordinate(aux)
   -- (aux|-cr.south) node[midway,right=1mm]{};
       \draw[spring] ([xshift=-5mm]g1.north) coordinate(aux)
   -- (aux|-cir.250) node[midway,right=1mm]{};
   \draw[spring] ([xshift=-10mm]g1.north) coordinate(aux)
   -- (aux|-cir.228) node[midway,right=1mm]{};
   \draw[spring] ([xshift=-15mm]g1.north) coordinate(aux)
   -- (aux|-cl.south) node[midway,right=1mm]{};
   \draw[spring2] ([xshift=-20mm]g1.north) coordinate(aux)
   -- (aux|-cl.south) node[midway](k1){};
   \draw[spring2] ([xshift=-25mm]g1.north) coordinate(aux)
   -- (aux|-cl.south) coordinate[midway](k2)node[midway,left=1mm]{$\Delta k_z$};
    \begin{scope}[on background layer]
    {
        \filldraw[lightgray](ft_sensor.west|- ft_sensor.south)--(ft_sensor.east|- ft_sensor.south) -- (cir.0)--(cir.180);
    }
    \end{scope}
    \draw[thin,dashed](k1|-cl.north) --++ (0,0.8)coordinate[pos=0.85](aus1);
    \draw[thin,dashed](k2|-cl.north) --++ (0,0.8);
    \draw[latex-latex](k1|-aus1)--(k2|-aus1) node[midway, above=1mm]{$\Delta x$};
    \node at (0,5) {Indenter};
    
    \end{tikzpicture}
    \caption{Spherical indenter inside the elastic foundation, $d$ is the penetration, $a$ is the projection of the surface of the circle in contact and $R_1$ is the equivalent radius of the 3D sphere.}
    \label{fig:contact}
\end{figure}
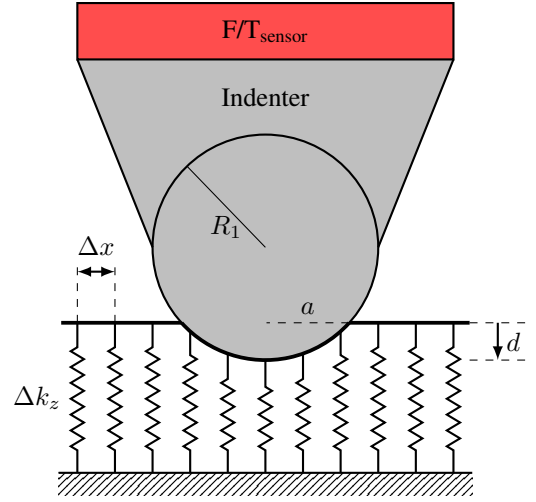

\noindent {\bf Spherical Indenter.}
Using the Hertzian theory, the profile of the indenter sphere can be
approximated with the 2D parabolic profile
$g(x) = \frac{x^2}{2 R_1}$~\cite{johnson1985contact}, where $R_1$ is
the radius of the sphere.  Let $d$ be the depth of penetration into
the material. The force exerted by an individual spring will vary
according to its position along the two-dimensional profile of the
indenter. For $\Delta x \rightarrow 0$, equation
\eqref{eq:single_spring} can be written as:
\begin{equation}
     f_{N}(x) = \frac{E_f}{1-\nu^2} \left( d - \frac{x^2}{2 R_1} \right).
     \label{eq:single_spring_inf}
\end{equation}
The part of the indenter in contact with the surface is
$a = \sqrt{2 R_1 d}$, hence to obtain the total normal force generated
by the material equation \eqref{eq:single_spring_inf} is integrated
from $-a$ to $a$ obtaining
\begin{equation}
  \begin{aligned}
    F_N(d) &=  \int_{-a}^{a} \left[ \frac{E_f}{1-\nu^2} \left( d -
          \frac{x^2}{2 R_1} \right) \right] dx = \\
    &=  \int_{-\sqrt{2 R_1 d}}^{\sqrt{2 R_1 d}}\left[
      \frac{E_f}{1-\nu^2} \left( d - \frac{x^2}{2 R_1} \right) \right]
      dx = \\
      &= \frac{4}{3} \frac{E_f}{1-\nu^2} d \sqrt{ 2 R_1 d} .
    \end{aligned}
    \label{eq:final_ext_contact}
\end{equation}
Notice that, imposing $R = 2R_1$, the same results of the Hertzian
theory can be derived. In Figure~\ref{fig:contact}, the 2D contact
model between a sphere and an elastic half-space is represented.

\noindent{\bf Generalisation to Symmetric indenters. }
The previous formulation can be extended to any axially symmetric
indenter. For simplicity, we define the elastic half-space at $z=0$,
thereby making $z$ the axis of symmetry. The surface of the indenter
is then parameterised by the $x$ and $y$ coordinates.

For a generic axial symmetric indenter, the profile can be written as
\begin{equation}
  \Tilde{z} = l_n(r) = c_n r^n ,
  \label{eq:3dprof}
\end{equation}
where $c_n$ is a constant depending on the shape of the profile,
$r = \sqrt{x^2+y^2}$ and $n$ is an arbitrary positive number. As
previously discussed, a two-dimensional profile and a linear elastic
foundation can model the contact between a three-dimensional object
and an elastic half-space. Therefore, profile~\eqref{eq:3dprof} is
defined in one dimension as
\begin{equation}
  \Tilde{z} = g_n(x) = \tilde{c}_n |x|^n ,
\end{equation}
where $\tilde{c}_n$ can be computed following the rule of
Hess~\cite{Popov2015MethodFriction}. Finally the normal force can be
written as
\begin{equation}
  F_N(d) = \frac{2 n}{n +1 } \frac{E_f}{1 - \nu^2} \Tilde{c}^{-\frac{1}{n}} d^{\frac{n+1}{n}}.
    \label{eq:f_gen}
\end{equation}
Using~\eqref{eq:f_gen}, it is possible to obtain the force equation
for the flat and for the parabolic indenter ($n=2$) as
\begin{align}
  & F_{N,flat}(d) = \frac{E_f}{1 - \nu^2} 2 a d, \\ 
  & F_{N,parab}(d) = \frac{4}{3} \frac{E_f}{1-\nu^2} d \sqrt{ R d}.
\end{align}

\subsection{Data Collection and Elasticity Estimation}
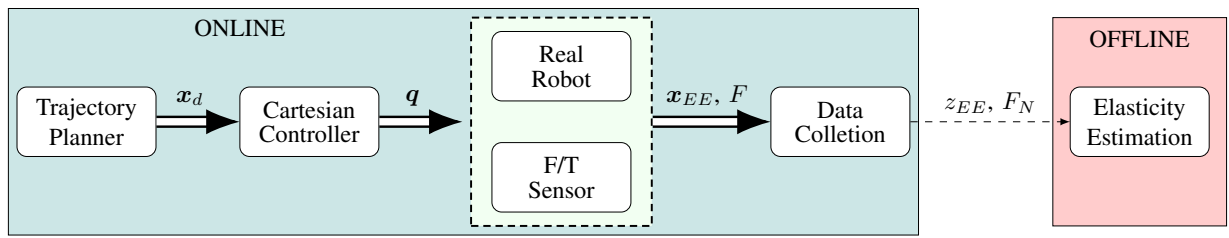
\begin{figure*}[t]
    \centering
    \usetikzlibrary{matrix,fit} 
\usetikzlibrary{calc,patterns,decorations.pathmorphing,decorations.markings,positioning,backgrounds,arrows.meta,shapes,fit,matrix}
    \tikzstyle{box} = [rectangle, rounded corners, minimum width=2cm, minimum height=1cm,text centered, draw=black, fill=white]
    \resizebox{0.9\textwidth}{!}{%
    \begin{tikzpicture}
        \node (box1) [box] {\shortstack{Trajectory \\ Planner} };
        \draw [line width=1pt, double distance=3pt,
             arrows = {-Latex[length=0pt 3 0.1]}] (box1.east) --++ (1.2,0) coordinate[pos=1](a1) node[pos=0.4,above,yshift=1mm]{$\boldsymbol{x}_d$};
        \node (box2) [box,right of=a1] {\shortstack{Cartesian \\ Controller} };
        \draw [line width=1pt, double distance=3pt,
             arrows = {-Latex[length=0pt 3 0.1]}] (box2.east) --++ (1.2,0) coordinate[pos=1.35](a2) node[pos=0.4,above,yshift=1mm]{$\boldsymbol{q}$};
        \node (cont) [draw, thick ,densely dashed, fill=green!5, minimum width=2.6cm, minimum height=3cm,right of=a2 ]{};
        \node (box3) [box, below of = cont,yshift=2mm]{\shortstack{F/T \\ Sensor}};
        \node (box4) [box, above of = cont,yshift=-2mm]{\shortstack{Real \\ Robot}};
        \draw [line width=1pt, double distance=3pt,
             arrows = {-Latex[length=0pt 3 0.1]}] (cont.east) --++ (1.7,0) coordinate[pos=1](a3) node[pos=0.4,above,yshift=1mm,xshift=1mm]{$\boldsymbol{x}_{EE}$, $F$};
        \node (box5) [box, right of=a3]{\shortstack{Data \\ Colletion}};
        \begin{pgfonlayer}{background}
            \node[draw,fill=teal!20,fit=(box1) (cont) (box5)] (cont2) {};
        \end{pgfonlayer}
        \node[above of= a1,yshift=0.35cm](text1){ONLINE};
        \draw[-latex,dashed] (box5.east) -- ++ (2.3,0) coordinate[pos=1](a4) node[pos=0.5,above]{$z_{EE}$, $F_N$};
        \node[box,right of= a4](box6){\shortstack{Elasticity \\ Estimation} };
        \node[above of= box6,yshift=0.2cm](text2){OFFLINE};
        \begin{pgfonlayer}{background}
            \node[draw,fill=red!20,fit=(box6) ,minimum height=3cm, minimum width=2.5cm] (cont3) {};
        \end{pgfonlayer}
    \end{tikzpicture}
}
    \caption{Estimation algorithm scheme illustrating the measurement
      acquisition procedure, function of the robot trajectory, and
      resulting in the estimation of the elasticity value.}
    \label{fig:schematic}
\end{figure*}

Position data and force data are collected from two different
sources. Since joint data are collected by the robotic arm encoders
and following the Denavit-Hartenberg
rule~\cite{Uigker1964AnMechanisms}, a homogeneous transformation
matrix corresponding to each of the $n$ joints can be built as
\begin{equation}
  \label{eq:RigidTransf}
  \boldsymbol{T}_i = 
  \begin{bmatrix}
    \boldsymbol{R}_i & \boldsymbol{p}_i \\
    \boldsymbol{0} & 1
  \end{bmatrix} ,
\end{equation} 
where $i$ is the joint number,
$\boldsymbol{R}_i \in \mathbb{R}^{3\text{x}3}$ is the orientation
matrix representing the orientation of the joint $i$ and
$\boldsymbol{p}_i \in \mathbb{R}^{3\text{x}1}$ is the position of the
joint $i$. The pose of the end-effector for an $m$ joint arm can then
be expressed as the product of the $m$ transformations
in~\eqref{eq:RigidTransf}, i.e.
\begin{equation}
  \label{eq:RigidEE}
  \boldsymbol{T}_{ee} = \boldsymbol{T}_1 \boldsymbol{T}_2 \cdots
  \boldsymbol{T}_m .
\end{equation}
The force data are collected directly from the force torque sensor.
To ensure synchronisation, the data are collected every control cycle
by reading the latest update. The control cycle operates with a
refresh rate that matches the slower sensor. Once the data have been
collected, they are processed offline. The flow of the procedure can
be seen in Figure~\ref{fig:schematic}.
As aforementioned, the data on the position of the end-effector are
obtained from the joint values. The penetration inside the body can be
computed using as reference the position where we start recording
positive forces. The force generated by the material is instead equal
to the force registered along the $z$-axis of the force torque sensor.

A least squares algorithm is used to estimate the best elasticity
value by minimising
\begin{equation}
  \mathcal{L} = \sum_{j=i}^{L} \left(F_{sensor,j} - \kappa ( d )^{n} \right)^2,
  \label{eq:ls}
\end{equation}
where $j$ is the number of acquired measurement results, $d$ is the
penetration, $n$ is the shape-factor discussed previously, and
$\kappa$ is related to the elasticity as a function of the indenter
shape. For example, using the sphere, $n=\frac{3}{2}$ and
\begin{equation}
    E_f = \kappa  \frac{3 ( 1 - \nu^2) }{ 4 \sqrt{R}} .
\end{equation}
Once challenge for the problem at hand is related to the difficulties
to locate the precise position of the specimen surface within the
robot framework, since the detection of the contact force is affected
by uncertainties. Therefore, a constrained minimisation problem is
defined on~\ref{eq:ls} as
\begin{equation}
    \begin{aligned}
    \min_{z_{surf}} \quad & \mathcal{L}\\
    \textrm{s.t.} \quad & -F_{unc} \leq F(z_{surf})\leq F_{unc}.
    \end{aligned}
    \label{eq:minprob}
\end{equation}
where $z_{surf}$ and $z_{EE}$ are the $z$-axis positions of the
surface and of the end-effector, respectively, $d = z_{surf} - z_{EE}$
by definition, $F_{unc}$ is the rated uncertainty of the sensor and
$F$ is the collected force.  This way, it is possible to find the
position of the surface with precision despite the uncertainties of
the force sensor.

\section{Experimental results}
\label{sec:experiments}
\begin{figure}[t]
  \centering
  \includegraphics[angle=-90,width=4.5cm]{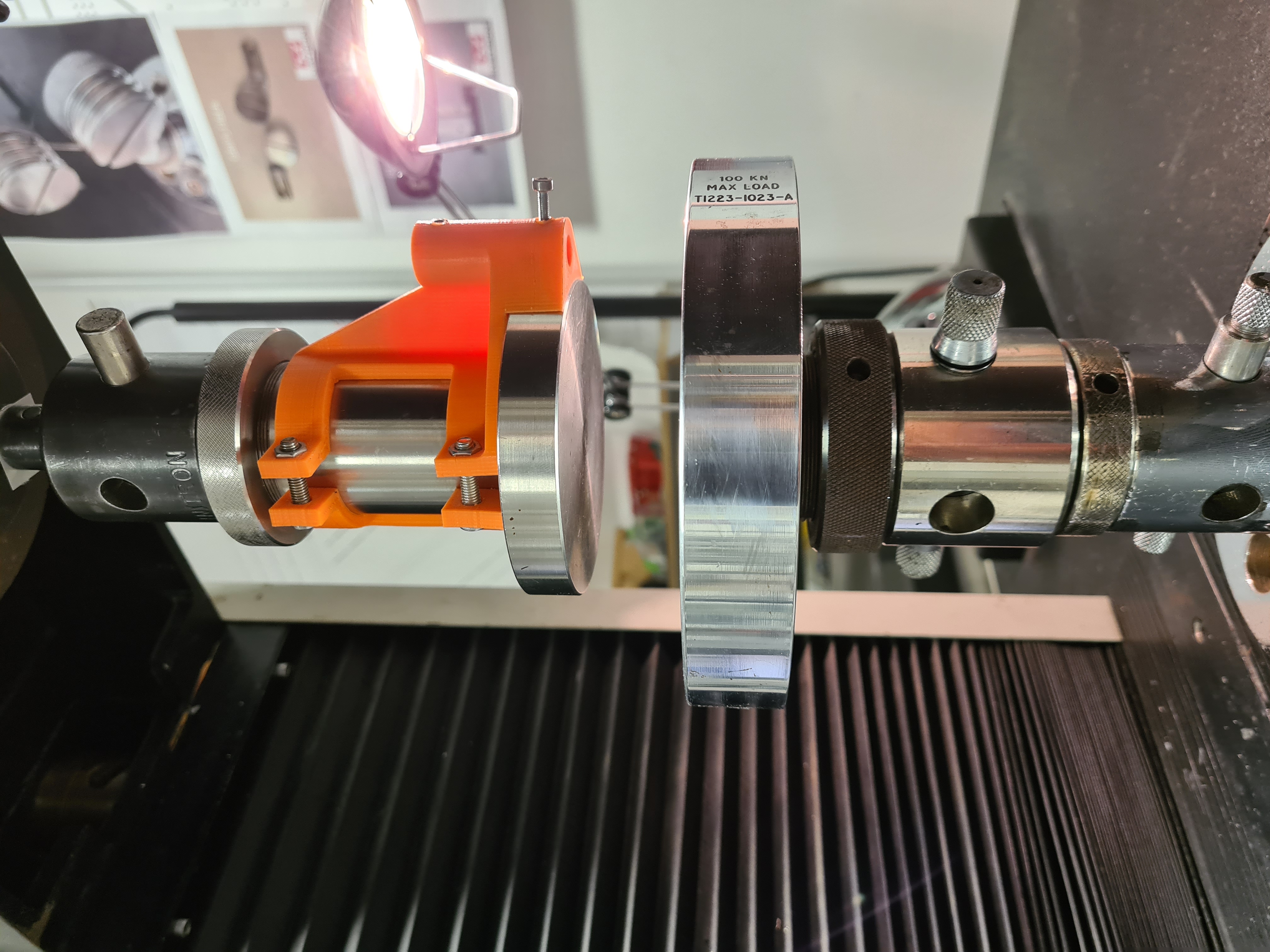}
  \caption{Instron® 4502 dynamometer used for the ground truth test of the specimens.}
  \label{fig:dyn}
\end{figure}

\begin{figure}[t]
    \centering
    \includegraphics[width=0.6\columnwidth]{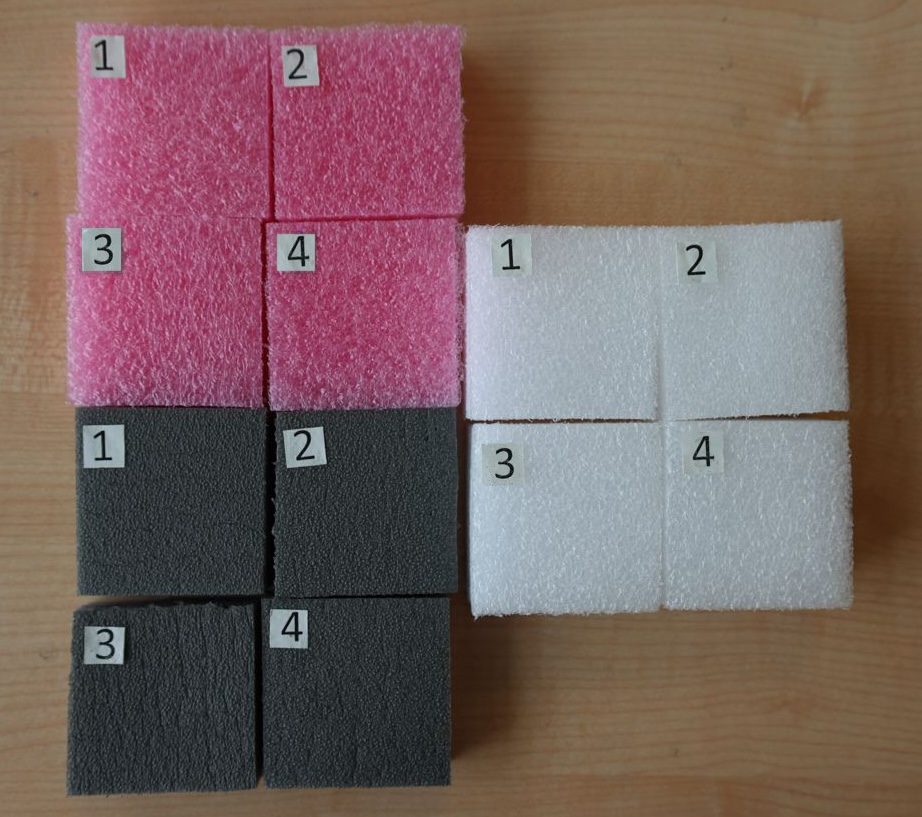}
    \caption{Specimens used during the experiments made of multilayer polymeric foams.}
    \label{fig:cubetti}
\end{figure}

The characterisation of the measurement system has been carried out
using a set of specimens that were tested with both a precise
measurement system used for compression testing and with the robotic
indenter. Quasi-static compression tests were performed at $23^\circ$C
using an Instron® 4502 dynamometer (Norwood, MA, USA) equipped with a
$\SI{50}{\kilo \newton}$ load cell, shown in Figure~\ref{fig:dyn}.
The tests were performed on parallelepiped-shaped specimens (see
Figure~\ref{fig:cubetti}) with $50\times50$~mm section and thickness
in range $20-30$~mm, at a cross-head speed of $50$~mm/min with
sampling rate of $25$~pt/s (or $25$~Hz) up to $5$~mm of compression.
The samples used in the experiments were polymeric foams made of
polyethylene and formed from multiple layers to achieve the desired
thickness for the experiments, which present elasticity values comparable with biological tissues~\cite{wells2011medical}. The geometric and physical
characteristics are specified in Table~\ref{tab:cube_param}.
\begin{table}[t]
    \centering
    \caption{Geometric and physical parameters of the specimens.}
    \label{tab:cube_param}
    \begin{tabular}{ccc}
    
    \textbf{Specimen} & \textbf{Density} [$\SI{}{\gram / \centi\meter^3}$] & \textbf{Length x Width x Height} [$\SI{}{\milli\meter^3}$]\\

    \textbf{White}   & $0.0350$ & $50 \times 50 \times 30$ \\
    
    \textbf{Pink}   & $0.0181$ & $50 \times 50 \times 20$ \\
    
      \textbf{Grey}   & $0.0236$ & $50 \times 50 \times 20$ \\
    
    \end{tabular}

\end{table}

\subsection{Robotic system}

The robot that has been used for the experiment is an Ur3e, a widely
used collaborative industrial robot. It has a maximum payload of
$3$~kg and a pose repeatability per ISO 9283 of $\pm 0.03$~mm. Since
the Ur3e is a $6$ degree-of-freedom robotic arm, we have $m = 6$
in~\eqref{eq:RigidEE}. At the end-effector of the manipulator is
attached a $6$-axis force-torque sensor, i.e., the BOTA System
SensONE. The sensor works at $800$~Hz with a peak-to-peak standard
uncertainty of $48\cdot 10^{-3}$~N in the $z$-direction. Attached to
the sensor there is a 3D printed indenter, whose position is computed
using the $m$ joint encoders measurements. The robotic arm is
controlled by a simple motion controller in the Cartesian space. This
means that given the Cartesian position of the end-effector as input,
the values of the joints are calculated by the robot inverse
kinematics. The convergence speed is controlled by a PD controller,
were the proportional gain was empirically set to $20$, while the
derivative gain to $0.5$, thus ensuring an end-effector trajectory
following with negligible errors.

\subsection{Results and analysis}

\begin{figure}[t]
  \centering
  \includegraphics[width=0.7\columnwidth]{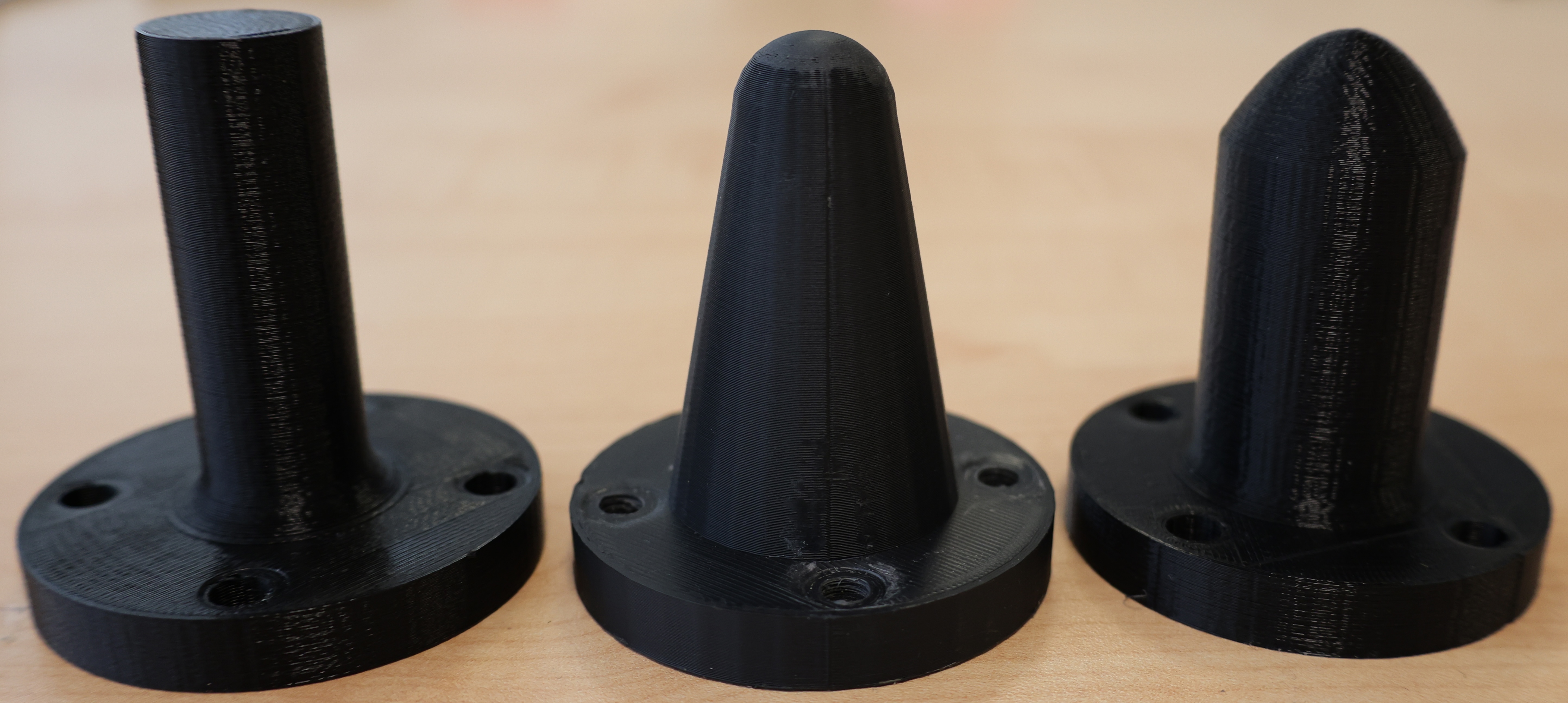}
  \caption{3D-printed tips utilised in the experiments. From left: the flat, the spherical and the parabolic tips.}
  \label{fig:tips}
\end{figure}

The tests carried out with the laboratory instrument in
Figure~\ref{fig:cubetti}, treated here as ground truth, have been
repeated using the Ur3e manipulator. The experiments were conducted
using three different tips: a flat tip with $1$~cm radius, a spherical
tip with $1$~cm radius and a parabolic tip with $1.17$~cm radius, all
depicted in Figure~\ref{fig:tips}.
For comparisons, the penetration velocity remained consistent at
$50$~mm per minute, as in the laboratory test. The indentation depth
is adjusted to $10\%$ of the sample thickness to remain within the
range of material
linearity~\cite{Dimitriadis2002DeterminationMicroscope,
  Qiang2011EstimatingThickness}, since any deeper penetration could
amplify the non-linear characteristics of the material that are not
modelled. During the experiment, the position and force in the
$z$-direction are collected for the following offline post-processing,
as described in Figure~\ref{fig:schematic}. Using~\eqref{eq:minprob},
we were able to reconstruct the force quite precisely with a mean
residual value that never exceeded $\SI{0.03}{\kilo \pascal^2}$ for
the spherical and paraboloidal tip and $\SI{0.06}{\kilo\pascal^2}$ for
the flat tip, as can be seen in Table~\ref{tab:res_exp}.
\begin{table}[t]
    \centering
    \caption{Residual resulting from the force reconstruction $\left[\SI{}{\kilo \pascal^2}\right]$.}
    \label{tab:res_exp}
    \begin{tabular}{cccc}
    
    \textbf{Specimen} & \textbf{Flat Tip} & \textbf{Sphere Tip} & \textbf{Paraboloid Tip} \\

    \textbf{White} & 0.0316 & 0.0240 &  0.0292  \\
    
    \textbf{Pink} & 0.0225 & 0.0250 & 0.0291 \\
    
    \textbf{Grey} & 0.0613 & 0.0208 & 0.0308 \\
                    
    \end{tabular}

\end{table}
The position of the surface for each specimen was estimated with
relatively high accuracy except for the flat tip, resulting in a mean
error for the spherical and paraboloidal tip of $0.38$~mm and for
$0.39$~mm, respectively. Instead, since the flat tip had non-linearity
of the phase after contact, we used the force measurements after
$20\%$ of penetration to estimate the specimen elasticity. The
estimated elasticity quantities are consistent with the ground truth,
as depicted in Table~\ref{tab:elast_table}.
\begin{table*}[t]
    \centering
    \caption{Results of the elastic value estimations for three
      different specimens using the three tips represented with the
      elasticity estimates $E$, the standard uncertainty $\sigma$, and
      the relative error $err$ with respect to the ground truth.}
    \label{tab:elast_table}
    \usetikzlibrary{matrix,fit} 

\definecolor{airforceblue}{rgb}{0.36,
0.54, 0.66}
\definecolor{cyan}{rgb}{0.3010, 0.7450, 0.9330}
\definecolor{orange}{rgb}{0.8500, 0.3250, 0.0980}
\definecolor{magenta}{rgb}{0.4940, 0.1840, 0.5560}
\begin{tikzpicture}[
  font=\sffamily,
  head color/.style args={#1/#2}{
    row 1 column #1/.append style={nodes={fill=#2}}},
]

\matrix[
   matrix of nodes, nodes in empty cells,
   nodes={minimum width=1.4cm, align=center,
          minimum height=2em, anchor=center},
   every even row/.style={nodes={fill=airforceblue!50}},
   column 1/.style= {nodes={fill=airforceblue, inner ysep=0}},
   row 1/.style= {nodes={text depth=0.2ex, text=black}},
   row 1 column 1/.style={nodes={fill=none, draw=none}},
   head color/.list={2/teal!70,3/teal!70,4/cyan!70,5/cyan!70,6/cyan!70,7/orange!70,8/orange!70,9/orange!70,10/magenta!70,11/magenta!70,12/magenta!70} 
  ] (m)
  {

      & $E$ $[\SI{}{\kilo\pascal}]$ & $\sigma$ $[\SI{}{\kilo\pascal}]$ & $E$ $[\SI{}{\kilo\pascal}]$ & $\sigma$ $[\SI{}{\kilo\pascal}]$ & $err$ $[\%]$ & $E$ $[\SI{}{\kilo\pascal}]$ & $\sigma$ $[\SI{}{\kilo\pascal}]$ & $err$ $[\%]$& $E$ $[\SI{}{\kilo\pascal}]$ & $\sigma$ $[\SI{}{\kilo\pascal}]$ & $err$ $[\%]$ \\
      White & $111$ & $13$   &   $119.6$   & $10.83$ & $-7.74$ & $123.75$   &  $7.39$    & $-10.81$  & $104.84$ & $6.33$ & $5.54$\\
      Pink & $136$ & $14$   & $154.49$     & $9.35$ & $-13.59$ & $144.03$   & $12.66$     & $-4.12$  & $123.21$ & $10.84$&  $9.40$\\
      Grey & $194$ & $17$   & $227.46$     & $7.35$ & $-17.24$ & $202.16$   & $20.29$     & $-5.88$  & $173.97$ & $19.61$ & $10.32$ \\
      };

    \node[text depth=0.2ex, text=black, fill= teal!70, anchor=south west,minimum width=2.8cm, align=center, minimum height=2em] at (m-1-2.north west) {Ground Truth};
    \node[text depth=0.2ex, text=black, fill= cyan!70, anchor=south west,minimum width=4.2cm, align=center, minimum height=2em] at (m-1-4.north west) {Flat Tip};
    \node[text depth=0.2ex, text=black, fill= orange!70, anchor=south west,minimum width=4.2cm, align=center, minimum height=2em] at (m-1-7.north west) {Spherical Tip};
    \node[text depth=0.2ex, text=black, fill= magenta!70, anchor=south west,minimum width=4.2cm, align=center, minimum height=2em] at (m-1-10.north west) {Paraboidal Tip};
\end{tikzpicture}
\end{table*}
The flat tip exhibits the poorest behaviour due to the previously
mentioned issues, leading to a maximum relative error of nearly
$20\%$. On the other hand, the estimates for the spherical and the
paraboloidal tip have a maximum relative error of $10\%$. Notice that
the standard uncertainty of the ground truth is almost entirely due to
the slight differences of the tested samples of the same material: a
variability of up to $10\%$ may be expected due to the intricate
structure of the expanded foam
characteristics~\cite{Weienborn2016DeformationMethods}, since the
average pore size significantly affects the mechanical properties and
can differ among specimens. However, an uncertainty of $10\%$ is more
than acceptable for medical applications given that cancers can be up
to $100$ times stiffer than normal soft
tissue~\cite{Nadan2007EUSLesions}.

\begin{figure}[t]
    \centering
    \includegraphics{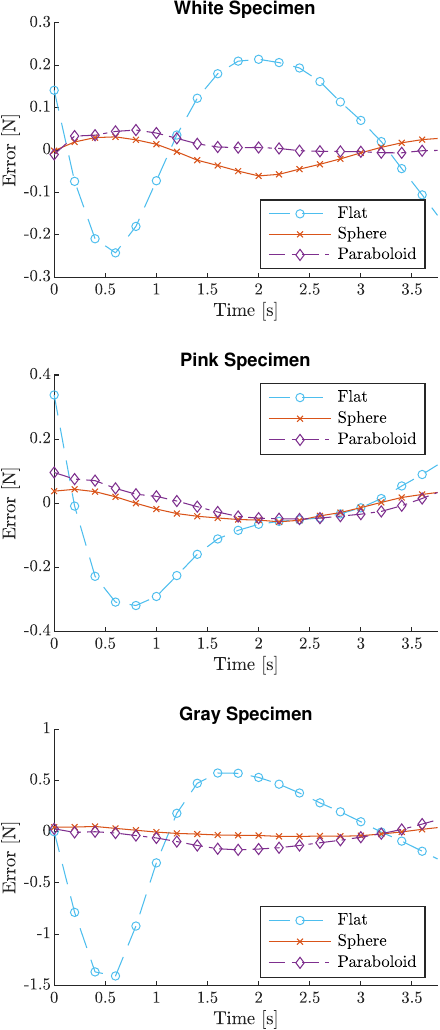}
    \caption{Average force error between the measured and the estimated value in the three specimens. In all experiments, the average error using the spherical and the paraboloid indenter is bounded, demonstrating the suitability of the model with these indenter shapes. On the other hand, the oscillating average error of the flat indenter does not ensure precise force estimation. 
      }
    \label{fig:forces}
\end{figure}

The error between the ground truth forces and the one derived from the
adopted models are reported in Figure~\ref{fig:forces}.
Again, the spherical and paraboloidal tips accurately reconstruct the
forces along the contact, while the model with a flat surface shows
less accuracy that maybe requires a nonlinear model to accurately
describe the contact between the flat tip and the specimens.

\begin{figure}[t]
    \centering
    \includegraphics{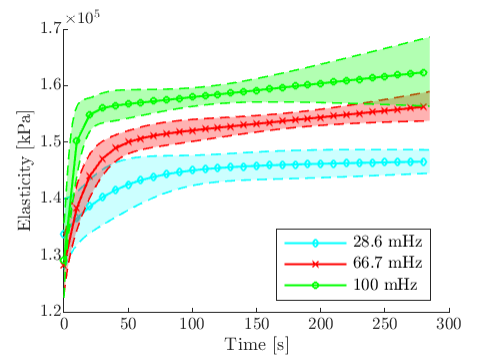}
    \caption{Change in stiffness when multiple experiments are performed without letting the specimen recover its characteristics. Each curve has its confidence interval of $95\%$.}
    \label{fig:mutli_palp}
\end{figure}

As materials such as expanded foams require time to recover their
characteristics after a test, we aim to investigate whether multiple
measurements can be taken in a short time frame while accurately
estimating the elasticity value. So we tried to understand if the
increase in the material stiffness that occurs in foams when they are
not allowed to restore to their initial properties after excitation
was measurable by our system. Figure~\ref{fig:mutli_palp} illustrates
how the elasticity changes with an increase in the
palpation/excitation frequency.
When excitations had higher frequencies (i.e., there is less time for
recovery from the previous deformation), higher measured elasticities
are observed. This phenomenon can be mathematically modelled as a sum
of two exponential functions:
\begin{equation}
  \label{eq:ElasticityModel}
  E(t) = c_1 e^{c_2 t} + c_3 e^{c_4 t} ,
\end{equation}
where $t$ us the time passed after the excitation, $c_i$ are tuning
constants and $E(t)$ the elasticity. It can be seen from
Figure~\ref{fig:mutli_palp} that the designed instrument is able to
also detect this subtle changes in the material elasticity, thus it
can be applied to estimate the parameters
of~\eqref{eq:ElasticityModel} and, more importantly, further proving
that our system is suitable for the foreseen medical applications.

\section{Conclusion}
\label{sec:conclusion}

In this paper, we have an experimental setup, based on the use of a
robotic device, to estimate the elastic parameter of a material. This
is fundamental to the development of robotic medical applications
based on the physical interaction between a probe and the patient's
body. The paper shows that it is possible to reconstruct the contact
force with high precision using different types of tips. By leveraging
these results, we could estimate the elasticity value with an accuracy
that is sufficient to detect imperfections in the human tissues that
could reveal the presence of possible diseases. 

Further work is required to improve the method. It is not reasonable
to assume that all actions performed by the robot will be quasi-static
and, hence, it will be necessary to implement a component that
considers the dynamic properties of the material. In addition, an
online estimation algorithm would be invaluable for automated
diagnostic applications based on precise tactile inspections of some
parts of the human body. In this paper, we have set the basis for this
future development showing the high degree of accuracy that these
measurements can potentially reach.

\section*{Acknowledgment}
We acknowledge the support of the MUR PNRR project FAIR - Future AI Research (PE00000013).




\bibliographystyle{IEEEtran}
\bibliography{references.bib}
%



\end{document}